\PassOptionsToPackage{table}{xcolor}

\documentclass{article}

\usepackage{microtype}
\usepackage{graphicx}
\usepackage{subcaption}
\usepackage{booktabs} 

\usepackage{hyperref}

\usepackage[preprint]{icml2026}

\usepackage{amsmath}

\usepackage{amssymb}
\usepackage{mathtools}
\usepackage{amsthm}
\usepackage{enumitem}

\usepackage{siunitx}
\usepackage[table]{xcolor}
\usepackage{multirow}
\usepackage{pifont}
\definecolor{rowblue}{RGB}{230,245,255}
\definecolor{darkgreen}{RGB}{0,150,0}
\newcommand{\cmark}{\ding{51}}
\newcommand{\xmark}{\ding{55}}

\usepackage[capitalize,noabbrev]{cleveref}

\theoremstyle{plain}

\theoremstyle{definition}

\theoremstyle{remark}

\usepackage{booktabs}
\usepackage[table]{xcolor}
\usepackage{tabularx}
\usepackage{array}
\usepackage{pifont}

\usepackage[textsize=tiny]{todonotes}

\icmltitlerunning{S2GS: Streaming Semantic Gaussian Splatting for Online Scene Understanding and Reconstruction}

\begin{document}

\twocolumn[
  \icmltitle{S2GS: Streaming Semantic Gaussian Splatting for Online Scene Understanding and Reconstruction}



  \icmlsetsymbol{equal}{*}

  \begin{icmlauthorlist}
    \icmlauthor{Renhe Zhang}{yyy}
    \icmlauthor{Yuyang Tan}{yyy}
    \icmlauthor{Jingyu Gong}{yyy}
    \icmlauthor{Zhizhong Zhang}{yyy}
    \icmlauthor{Lizhuang Ma}{yyy}
    \icmlauthor{Yuan Xie}{yyy}
    \icmlauthor{Xin Tan}{yyy,comp}

  \end{icmlauthorlist}

  \icmlaffiliation{yyy}{East China Normal University}
  \icmlaffiliation{comp}{Shanghai Artificial Intelligence Laboratory}

  \icmlcorrespondingauthor{Xin Tan}{xtan@cs.ecnu.edu.cn}

  \icmlkeywords{}
  \vskip 0.3in
]



\printAffiliationsAndNotice{}  

\begin{abstract}
Existing offline feed-forward methods for joint scene understanding and reconstruction on long image streams often repeatedly perform global computation over an ever-growing set of past observations, causing runtime and GPU memory to increase rapidly with sequence length and limiting scalability. We propose Streaming Semantic Gaussian Splatting (S2GS), a strictly causal, incremental 3D Gaussian semantic field framework: it does not leverage future frames and continuously updates scene geometry, appearance, and instance-level semantics without reprocessing historical frames, enabling scalable online joint reconstruction and understanding. S2GS adopts a geometry–semantic decoupled dual-backbone design: the geometry branch performs causal modeling to drive incremental Gaussian updates, while the semantic branch leverages a 2D foundation vision model and a query-driven decoder to predict segmentation masks and identity embeddings, further stabilized by query-level contrastive alignment and lightweight online association with an instance memory. Experiments show that S2GS matches or outperforms strong offline baselines on joint reconstruction-and-understanding benchmarks, while significantly improving long-horizon scalability: it processes 1,000+ frames with much slower growth in runtime and GPU memory, whereas offline global-processing baselines typically run out of memory at around 80 frames under the same setting.

\end{abstract}

\section{Introduction}

\begin{figure}[ht]
  \centering 
  \includegraphics[width=\columnwidth]{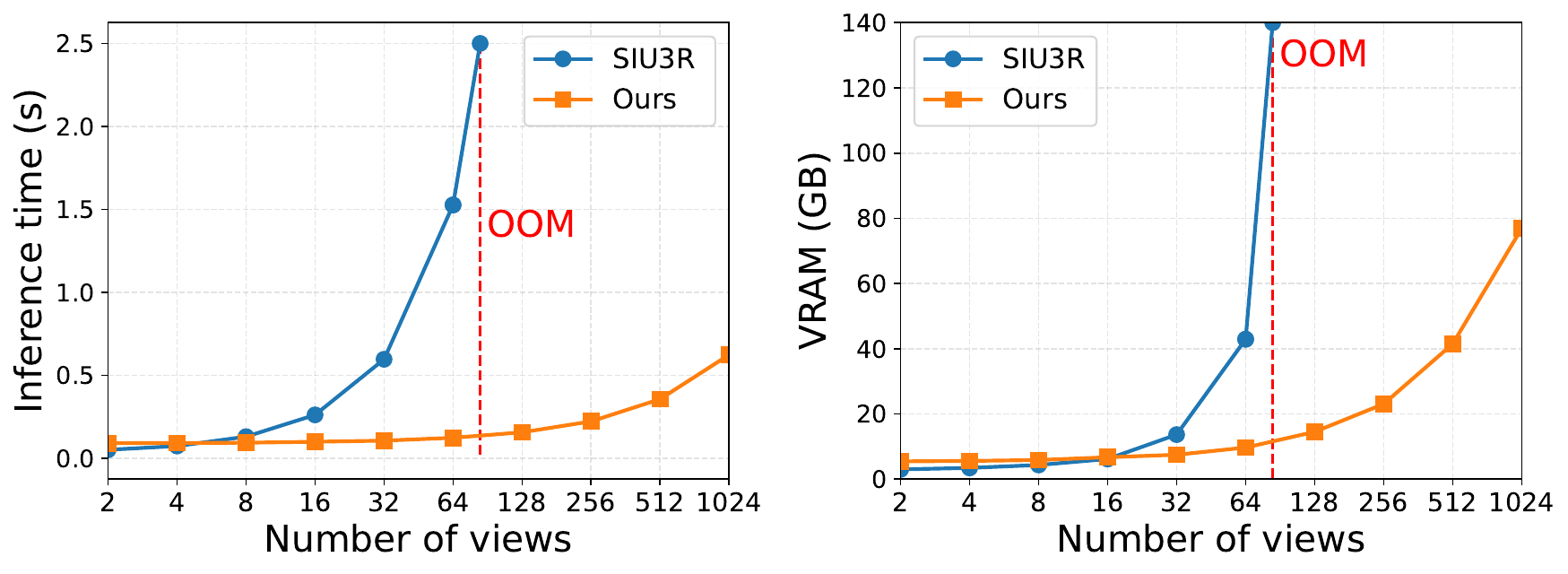}
  \caption{
    Comparison of current-frame inference time and GPU memory usage between S2GS (Ours) and the recent advanced joint reconstruction and understanding method, SIU3R\cite{siu3r}, under varying sequence lengths in the online setting.
  }
  \label{fig1}
\end{figure}

Recently, feed-forward methods~\cite{siu3r,uni3r,Uniforward} built upon 3D Gaussian Splatting (3DGS)~\cite{3dgs} have made substantial progress in jointly modeling geometry, appearance, and semantics for 3D reconstruction and scene understanding, enabling downstream applications such as robotic perception~\cite{app1}, AR/VR~\cite{app2}, and digital twins~\cite{app3}. 
However, most existing approaches remain offline-global in the sense that, as new frames arrive, they repeatedly recompute cross-frame interactions over the growing history. 
While effective for short sequences, this paradigm scales poorly: both runtime and memory typically grow rapidly with the number of views, hindering long-horizon online scenarios. 
As shown in Figure~\ref{fig1}, even on an H200 GPU equipped with 140 GB of VRAM, SIU3R \cite{siu3r} still encounters an out-of-memory (OOM) after processing approximately 80 frames, exposing a fundamental limitation of current joint modeling paradigms under long input streams. This phenomenon indicates that, for long-running online systems, there is an urgent need for an incremental modeling approach that does not require repeatedly reprocessing historical frames.

Meanwhile, recent advances \cite{streamgs,longsplat, cut3r,stream3r,streamvggt, infinitevggt} in streaming reconstruction have demonstrated better time and memory scalability. However, most existing approaches remain limited to streaming modeling of geometry and appearance, lacking semantic scene understanding and instance-level, decomposable representations, and thus falling short for downstream applications that require both reconstruction and understanding. 
More fundamentally, in real-world online scenarios, inputs arrive sequentially over time and the system must update its state and produce outputs on the fly. This naturally imposes a causal constraint on online joint reconstruction and understanding: at each time step, the model can only rely on the current observation and a persistent state accumulated from the past, without access to future information or global corrections via reprocessing historical frames. Under this constraint, how to incorporate stable and temporally consistent semantic understanding while preserving the scalability of streaming inference remains an open problem. 
Based on the above gaps, we revisit online joint 3D reconstruction and semantic understanding for long input streams, and propose \textbf{Streaming Semantic Gaussian Splatting (S2GS)}. S2GS is designed to match the operating characteristics of real-world online systems: without repeatedly reprocessing historical frames, it incrementally maintains scene geometry, appearance, and an instance-aware semantic field, thereby unifying streaming reconstruction and semantic understanding within a single framework.

S2GS addresses two core challenges in strictly causal online joint modeling: (i) maintaining stable geometry without future-view corrections, and (ii) preserving temporally consistent instance identities under view-dependent semantic observations. To this end, we adopt reprocessing-free streaming state updates and combine geometry--semantic decoupling with identity stabilization mechanisms tailored for long-horizon inference, thereby enabling online joint reconstruction and understanding. Our contributions are summarized as follows:
\begin{enumerate}[leftmargin=*, itemsep=1pt, parsep=0pt, topsep=2pt]
\item We propose S2GS, a strictly causal and reprocessing-free framework for online joint 3D reconstruction and scene understanding, which incrementally maintains scene geometry, appearance, and an instance-level semantic field.
\item We introduce a geometry–semantic decoupled dual-backbone design tailored for streaming inference: the geometry stream performs frame-wise causal aggregation under geometric priors for stable scene maintenance, while the semantic stream independently extracts per-frame multi-scale features using a 2D foundation model, preventing geometric update noise from corrupting semantic representations.
\item We propose streaming-specific semantic and identity stabilization mechanisms: during training, query-level contrastive alignment improves cross-frame consistency, and during inference, a lightweight instance-memory association reduces ID switches; additionally, we introduce a lightweight query semantic projector with distillation to align queries to a vision--language semantic space, enabling language-conditioned query retrieval and online open-vocabulary segmentation.
\item Across multiple joint reconstruction-and-understanding benchmarks and long-horizon online settings, S2GS achieves performance on par with or better than strong offline baselines, while significantly outperforming offline global paradigms in scalability with respect to sequence length, in terms of both inference runtime and GPU memory growth.
\end{enumerate}

\section{Related Work}

\textbf{Feed-Forward 3D Reconstruction and Simultaneous Understanding.} Scene-by-scene optimization–based reconstruction paradigms (e.g., NeRF \cite{nerf} and 3DGS \cite{3dgs}) can deliver high-quality rendering, but optimizing a single scene often takes minutes to hours, making them ill-suited for real-time applications. Recent feed-forward reconstruction approaches have substantially accelerated the reconstruction process: both Gaussian-based methods \cite{pixelsplat,mvsplat,noposplat,anysplat} and point-map–based methods \cite{dust3r,mast3r,vggt,fast3r} demonstrate improved efficiency. However, most of these works focus primarily on recovering geometry and appearance, lacking semantic embeddings for representation, which limits their ability to further understand and parse the content of 3D environments. More recent work has begun to jointly address scene understanding and 3D reconstruction. Uni3R \cite{uni3r} and Uniforward \cite{Uniforward} embed semantic features into Gaussian points, enabling zero-shot 3D semantic segmentation with arbitrary text prompts. SIU3R \cite{siu3r}, meanwhile, uses learnable queries to endow the model with 3D understanding capability.  IGGT \cite{iggt} proposes an end-to-end unified Transformer that learns 3D geometry and instance-level semantics from 2D images, enabling instance-aware 3D scene understanding. Despite their ability to predict geometry/appearance and semantic information simultaneously, these approaches largely follow an offline feed-forward paradigm: given an image sequence, they repeatedly encode, match, and globally optimize over the entire sequence (including historical frames) to obtain consistent reconstruction and semantic results. This paradigm is feasible for short to medium sequences, but its computation and memory costs often grow rapidly with the number of views, directly hindering scalability to long sequences, large-scale scenes, and online real-time settings.

\textbf{Online 3D Reconstruction of Image Streams.} The streaming paradigm offers a more scalable route for 3D reconstruction. In dense 3D geometric reconstruction, several representative works have advanced along this line: Spann3R \cite{spann3r} augments DUSt3R \cite{dust3r} with a memory-enhanced module to strengthen multi-view geometric fusion, while CUT3R \cite{cut3r} adopts an RNN-like \cite{rnn} incremental architecture that processes unstructured inputs step by step, enabling streaming inference. Going further, STream3R \cite{stream3r} and StreamVGGT \cite{streamvggt} formulate streaming 3D reconstruction as a decoder-only causal Transformer task and leverage KV caching and causal attention to achieve scalable online reconstruction over long image sequences. For novel view synthesis, StreamGS \cite{streamgs} proposes an online generalizable 3D GS reconstruction method for uncalibrated image streams, enabling efficient, continuous, high-fidelity reconstruction without per-scene optimization. Although the above methods have made notable progress in streaming geometric modeling or view synthesis, they mostly focus on geometry or appearance modeling itself and do not yet fully address the joint modeling of semantic information in streaming scene reconstruction. In contrast, we propose an incremental framework for joint scene and semantic reconstruction, which updates geometric structure and semantic representations online as new images arrive, without repeatedly performing forward passes over historical observations. This enables unified modeling of scene geometry and semantics while maintaining efficient streaming inference. In addition, online 3D reconstruction has been extensively studied in the SLAM literature \cite{niceslam,gaussianslam,gs3lam,omnimap,vtg}. Such methods typically rely on explicit pose estimation and graph optimization, and often assume additional sensing (e.g., depth/IMU) or strict camera calibration. In contrast, our method achieves synchronized online reconstruction of geometry and semantics using only uncalibrated and unposed RGB video stream.

\newcolumntype{C}{>{\centering\arraybackslash}p{0.8em}}
\begin{table}[t]
\centering
\caption{Comparison of S2GS with prior paradigms for 3D reconstruction and scene understanding.
\textbf{SC}: strictly causal; \textbf{RF}: reprocessing-free; \textbf{IS}: instance-level semantics; \textbf{SS}: streaming scalability; \textbf{RO}: RGB-only (no depth or additional sensors).}
\label{tab:paradigm}

\renewcommand{\arraystretch}{1.05}
\setlength{\tabcolsep}{6pt}

\begin{tabularx}{\columnwidth}{>{\raggedright\arraybackslash}X C C C C C}
\toprule
\textbf{Method Category} & \textbf{SC} & \textbf{RF} & \textbf{IS} & \textbf{SS} & \textbf{RO} \\
\midrule
Offline FF joint methods
& \textcolor{red}{\xmark} & \textcolor{red}{\xmark} & \textcolor{darkgreen}{\cmark} & \textcolor{red}{\xmark} & \textcolor{darkgreen}{\cmark} \\
Streaming point recon.\ only
& \textcolor{darkgreen}{\cmark} & \textcolor{darkgreen}{\cmark} & \textcolor{red}{\xmark} & \textcolor{darkgreen}{\cmark} & \textcolor{darkgreen}{\cmark} \\
Streaming GS recon.\ only
& \textcolor{darkgreen}{\cmark} & \textcolor{darkgreen}{\cmark} & \textcolor{red}{\xmark} & \textcolor{darkgreen}{\cmark} & \textcolor{darkgreen}{\cmark} \\
SLAM GS joint methods
& \textcolor{darkgreen}{\cmark} & \textcolor{darkgreen}{\cmark} & \textcolor{darkgreen}{\cmark} & \textcolor{darkgreen}{\cmark} & \textcolor{red}{\xmark} \\
\rowcolor{rowblue}
S2GS (Ours)
& \textcolor{darkgreen}{\cmark} & \textcolor{darkgreen}{\cmark} & \textcolor{darkgreen}{\cmark} & \textcolor{darkgreen}{\cmark} & \textcolor{darkgreen}{\cmark} \\
\bottomrule
\end{tabularx}
\end{table}

\textbf{Paradigm Comparison and Positioning.} The above lines of work differ mainly in whether they support (i) strictly causal inference, (ii) reprocessing-free streaming updates without reprocessing history, and (iii) instance-level semantic understanding under scalable long-horizon settings.
To make these capability trade-offs explicit, Table~\ref{tab:paradigm} summarizes representative paradigms from this perspective.
S2GS is positioned to combine the desirable properties: it performs strictly causal, reprocessing-free online updates from RGB-only uncalibrated streams, while maintaining instance-level semantics with scalable long-horizon inference.

\section{Method}

\begin{figure*}[t]
  \centering
  \includegraphics[width=0.95\textwidth]{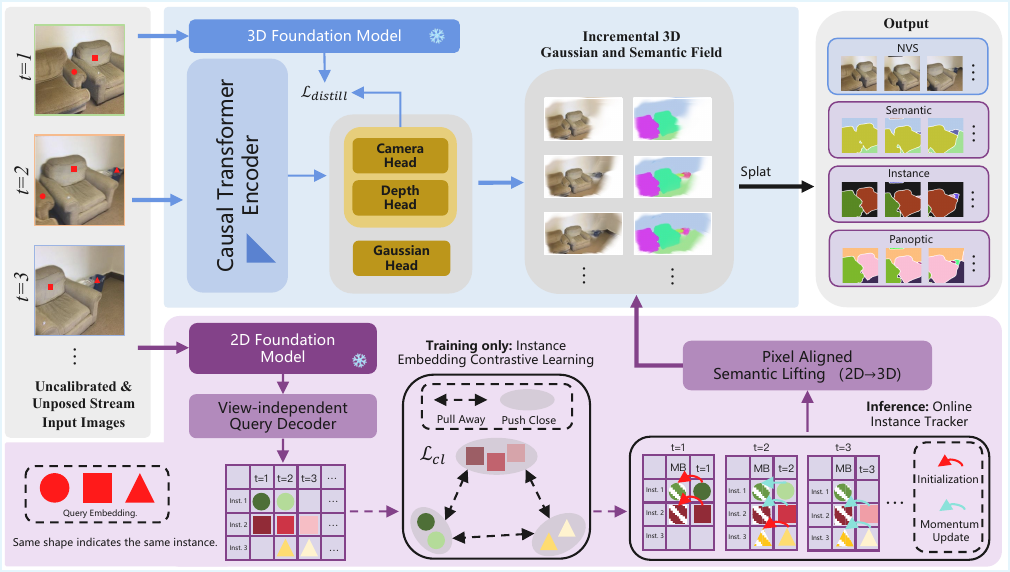}
  \caption{\textbf{Overview of S2GS.}
S2GS processes an uncalibrated and unposed RGB image stream in a strictly causal manner.
A causal Transformer encoder, guided by geometric priors from a 3D foundation model, predicts camera parameters, depth, and Gaussian attributes to incrementally construct 3D Gaussian representation. A decoupled semantic stream leverages a 2D foundation model and a query-driven decoder to produce per-view semantic and instance predictions.
Query-level contrastive learning and an online instance memory bank (MB) stabilize instance identities over time.
Semantic confidence is lifted to the 3D Gaussian field and decoded via splatting, enabling unified novel view synthesis, semantic segmentation, instance segmentation, and panoptic segmentation without revisiting past frames.}
  \label{fig:pipeline}
\end{figure*}

\subsection{Overview and Online Setting}

We consider online 3D reconstruction and instance-level semantic understanding from an uncalibrated RGB video stream
$\mathcal{I}=\{I_t\}_{t=1}^{T}$, where $I_t\in\mathbb{R}^{3\times H\times W}$.
S2GS operates under a strictly causal setting: at time $t$, the model processes only the current frame $I_t$ together with persistent states accumulated from previous steps, without reprocessing historical frames.
The model maintains a persistent 3D Gaussian scene representation and an instance-aware semantic state, enabling scalable long-horizon streaming inference.
As illustrated in Figure~\ref{fig:pipeline}, S2GS decouples geometry and semantics into two streams: a causal geometry stream for incremental reconstruction and a semantic stream for per-frame prediction and identity maintenance.

\subsection{Causal Transformer for 3D Gaussian Regression}

\textbf{Causal Transformer with online state.}
Following prior designs~\cite{vggt,stream3r}, each frame is encoded into visual tokens \cite{dinov2} and processed by a causal Transformer encoder.
Causality is enforced by a temporal attention mask that restricts tokens at time $t$ to attend only to the history prefix $\{1,\dots,t\}$:
\begin{equation}
M_{t,\tau}=
\begin{cases}
0, & \tau\le t,\\
-\infty, & \tau>t.
\end{cases}
\label{eq:causal_mask}
\end{equation}
This design allows parallel processing of training clips while remaining equivalent to an autoregressive causal model.
At inference, key/value tensors from past frames are cached and reused, enabling efficient long-horizon streaming without re-forwarding previous inputs.

\textbf{Incremental 3D Gaussian construction.}
Under the causal constraint, the Transformer aggregates information from $\{I_\tau\}_{\tau \le t}$ to form geometry features $H_t$.
We attach three lightweight heads to predict a dense depth map $\hat{D}_t$, camera parameters $\hat{P}_t$, and per-pixel Gaussian attributes $\hat{A}_t$.
To stabilize online reconstruction, we follow~\cite{uni3r,anysplat} and distill geometric supervision from a strictly causal pretrained 3D foundation model (teacher)~\cite{stream3r}.
Specifically, at each time step $t$, the teacher provides pseudo depth $\tilde{D}_t$ and pseudo camera/pose $\tilde{P}_t$.
We supervise the student by (i) an $\ell_2$ loss on depth over valid pixels and (ii) a Huber loss on the camera parameters, encouraging the predicted geometry (depth and camera) to remain consistent with the teacher.
With the distilled $\hat{D}_t$ and $\hat{P}_t$, we back-project pixels to obtain pixel-aligned 3D Gaussian centers, and combine them with the predicted attributes $\hat{A}_t$:

\begin{equation}
\mathcal{G}_t=\mathrm{BackProj}\!\left(\hat{D}_t,\hat{P}_t\right)\ \oplus\ \hat{A}_t.
\label{eq:gaussian_construct}
\end{equation}
The resulting Gaussians are incrementally accumulated into a persistent global scene representation and used for differentiable rendering.

\subsection{Online Instance Tracking and Semantic Stabilization}
\textbf{Decoupled Semantic Stream with Query-based Segmentation.}
To prevent interference from geometric updates, semantic representation learning is decoupled from geometric modeling.
Each incoming frame $I_t$ is encoded by a frozen 2D vision foundation model ~\cite{siglip2} to extract robust semantic features.
A lightweight adapter \cite{adapter} converts these features into multi-scale representations, which are consumed by a query-based mask-classification decoder~\cite{mask2former}.
At each time step, a fixed set of learnable queries attends to the current frame to produce per-frame masks, class scores, and query embeddings:
\begin{equation}
\mathbf{Q}_t = \mathrm{Dec}\!\left(\mathbf{Q}_0,\ \{\mathbf{F}_t^{(m)}\}\right).
\label{eq:query_decode_overview}
\end{equation}
The resulting query embeddings $\mathbf{Q}_t$ serve as identity descriptors for subsequent instance association.

\textbf{Online Instance Stabilization via Query-level Contrastive Learning.}
During training, we align predicted queries to ground-truth instances independently for each frame using Hungarian \cite{hm} matching and apply supervised contrastive learning to the aligned query embeddings.
This encourages embeddings corresponding to the same physical instance across frames to form compact clusters, while separating different instances.
Concretely, let $\mathbf{z}_i$ denote a normalized query embedding with instance label $y_i$, and let $\mathcal{P}(i)=\{p\neq i\mid y_p=y_i\}$ be the positive set for anchor $i$.
We optimize the supervised contrastive loss:
\begin{equation}
\mathcal{L}_{\mathrm{cl}}
=
\sum_{i=1}^{|\mathcal{Z}|}
\frac{-1}{|\mathcal{P}(i)|}
\sum_{p\in\mathcal{P}(i)}
\log
\frac{\exp(\mathbf{z}_i^\top \mathbf{z}_p / \tau)}
{\sum_{a\neq i}\exp(\mathbf{z}_i^\top \mathbf{z}_a / \tau)},
\label{eq:supcon}
\end{equation}
where $\tau$ is a temperature hyper-parameter and anchors with $|\mathcal{P}(i)|=0$ are ignored.

At inference, per-frame predictions are associated with a lightweight instance memory bank using cosine similarity and bipartite matching:
\begin{equation}
A_{i,k}^t = \cos(\mathbf{z}_i^t,\ \bar{\mathbf{z}}_k^{t-1}),
\label{eq:affinity}
\end{equation}
where $\mathbf{z}_i^t$ denotes a normalized query embedding and $\bar{\mathbf{z}}_k^{t-1}$ is the prototype embedding of an existing instance.
Matched prototypes are updated via exponential moving average:
\begin{equation}
\bar{\mathbf{z}}_k^{t}
=
\mathrm{Norm}\!\left((1-\alpha)\bar{\mathbf{z}}_k^{t-1} + \alpha\,\mathbf{z}_i^t\right),
\label{eq:ema_update}
\end{equation}
yielding temporally consistent instance identities under streaming inference.

\textbf{2D-to-3D Semantic Lifting.} Per-frame semantic and instance predictions are lifted into the 3D domain by assigning them to pixel-aligned 3D Gaussians.
Semantic attributes are treated analogously to appearance attributes and fused through differentiable Gaussian splatting, naturally aggregating multi-view evidence based on geometry and visibility.
The resulting 3D semantic field can be rendered from arbitrary viewpoints, producing temporally and view-consistent semantic and instance predictions.

\subsection{Language-driven Open-vocabulary Segmentation}
\label{sec:open_vocab}

Following SIU3R~\cite{siu3r}, we formulate language-driven segmentation as language-conditioned query retrieval to support open-vocabulary segmentation.
The key difference is that SIU3R~\cite{siu3r} operates in an offline multi-view setting, where queries are jointly updated via omnidirectional cross-view attention, implicitly aggregating observations of the same instance across viewpoints and yielding temporally consistent instance semantics without explicit cross-frame alignment or post-processing.
In contrast, we focus on streaming inference, where queries are generated per frame and updated online over time, making global multi-view interaction infeasible.
Motivated by this discrepancy, we design a streaming-oriented language-conditioned query retrieval mechanism that explicitly aligns the semantic space and stabilizes temporal dynamics, enabling robust language matching under continuously updated queries.

\textbf{Query-to-semantic projection.}
Query embeddings $\mathbf{Q}_t[n]$ from the semantic decoder are internal representations optimized for mask prediction and instance association, and are not guaranteed to lie in the joint vision--language semantic space of 2D foundation vision model \cite{siglip2}.
To bridge this gap, we introduce a lightweight Query Semantic Projector $g_\theta(\cdot)$ that maps each per-frame query embedding to the 2D foundation vision model \cite{siglip2} embedding space:
\begin{equation}
\mathbf{u}_{t,n} = g_\theta(\mathbf{Q}_t[n]) \in \mathbb{R}^{C_s},
\end{equation}
where $C_s$ denotes the SigLIP2 embedding dimension.
During training, we freeze the semantic decoder and optimize only the projector parameters $\theta$. To provide a semantic teacher signal, we apply the predicted mask $\mathbf{m}_{t,n}$ to the input image and encode the masked region using the frozen SigLIP2 image encoder,
yielding a normalized teacher embedding $\mathbf{v}_{t,n}$.
We then align the projected query embedding $\mathbf{u}_{t,n}$ with $\mathbf{v}_{t,n}$ using a cosine regression loss, so that projected queries become directly comparable to SigLIP2 text embeddings.

\textbf{Stabilizing distillation under momentum-updated queries.}
At inference, queries are updated across frames via momentum to improve instance-level stability, which induces temporal variations in query embeddings.
To make the projection robust to such dynamics, we enforce \emph{instance-level semantic invariance} during training:
supervised query-level contrastive learning encourages embeddings corresponding to the same physical instance across frames to form a compact cluster, and during distillation we randomly aggregate via weighted averaging same-instance queries from different views and distill the aggregated embedding toward the same SigLIP2 \cite{siglip2} teacher representation.
This explicitly accounts for embedding drift induced by momentum updates at test time.

\textbf{Language-driven query retrieval.}
At test time, given a text description $r$, we obtain a normalized text embedding $\mathbf{e}_r$ using the SigLIP2 \cite{siglip2} text encoder and compute cosine similarity with the projected queries,
\begin{equation}
s_{t,n} = \cos(\mathbf{u}_{t,n}, \mathbf{e}_r), \quad
n^\star = \arg\max_n s_{t,n}.
\end{equation}
We return the corresponding mask $\mathbf{m}_{t,n^\star}$ as the segmentation result.

\begin{table*}[t]
\centering
\caption{Comparison with feed-forward methods on the ScanNet \cite{scannet} dataset under short-sequence inputs.
``$\bullet$'', ``$\dagger$'', and ``$\star$'' denote reconstruction-only, understanding-only, and joint reconstruction-and-understanding methods, respectively.}
\label{tab:scannet_feedforward}

\renewcommand{\arraystretch}{1.05}
\setlength{\tabcolsep}{3pt}
\footnotesize

\begin{tabular}{
l
S[table-format=2.2]
S[table-format=1.3]
S[table-format=2.2]
S[table-format=2.2]
S[table-format=2.2]
S[table-format=2.2]
S[table-format=1.3]
S[table-format=2.2]
S[table-format=2.2]
S[table-format=2.2]
}
\toprule
\multirow{2}{*}{\textbf{Method}}
& \multicolumn{5}{c}{\textbf{2 views}}
& \multicolumn{5}{c}{\textbf{8 views}} \\
\cmidrule(lr){2-6}\cmidrule(lr){7-11}
& {\textbf{PSNR$\uparrow$}} & {\textbf{SSIM$\uparrow$}} & {\textbf{mIoU$\uparrow$}} & {\textbf{T-mIoU$\uparrow$}} & {\textbf{T-SR$\uparrow$}}
& {\textbf{PSNR$\uparrow$}} & {\textbf{SSIM$\uparrow$}} & {\textbf{mIoU$\uparrow$}} & {\textbf{T-mIoU$\uparrow$}} & {\textbf{T-SR$\uparrow$}} \\
\midrule
$\bullet$ pixelSplat~\cite{pixelsplat}
& 24.76 & 0.804 & {-} & {-} & {-}
& {-} & {-} & {-} & {-} & {-} \\
$\bullet$ MVSplat~\cite{mvsplat}
& 23.63 & 0.784 & {-} & {-} & {-}
& {-} & {-} & {-} & {-} & {-} \\
$\bullet$ NoPoSplat~\cite{noposplat}
& 25.27 & 0.811  & {-} & {-} & {-}
& {-} & {-} & {-} & {-} & {-} \\
$\dagger$ Mask2Former~\cite{mask2former}
& {-} & {-} & 47.32 & 42.11 & \underline{90.17}
& {-} & {-} & \underline{45.85} & \underline{31.93} & \underline{66.55} \\
$\dagger$ LSeg~\cite{lseg}
& {-} & {-} & 33.27 & {-} & {-}
& {-} & {-} & 32.87 & {-} & {-} \\
$\star$ LSM~\cite{lsm}
& 22.38 & 0.714 & 31.31 & {-} & {-}
& {-} & {-} & {-} & {-} & {-} \\
$\star$ Uni3R~\cite{uni3r}
& \underline{25.44} & \underline{0.812} & 32.18 & {-} & {-}
& 18.16 & 0.627 & 33.75 & {-} & {-} \\
$\star$ SIU3R~\cite{siu3r}
&  \textbf{25.79} &  \textbf{0.819} & \underline{47.83} & \underline{44.25} & 85.07
& \underline{19.74} & \underline{0.653} & 44.78 & 29.41 & 62.93 \\
\rowcolor{rowblue}
$\star$ S2GS (Ours)
& {24.90} & {0.810} &  \textbf{52.35} &  \textbf{44.89} &  \textbf{93.73}
&  \textbf{20.83} &  \textbf{0.685} &  \textbf{49.53} &  \textbf{33.34} &  \textbf{82.49} \\
\midrule
\multirow{2}{*}{\textbf{Method}}
& \multicolumn{5}{c}{\textbf{14 views}}
& \multicolumn{5}{c}{\textbf{32 views}} \\
\cmidrule(lr){2-6}\cmidrule(lr){7-11}
& {\textbf{PSNR$\uparrow$}} & {\textbf{SSIM$\uparrow$}} & {\textbf{mIoU$\uparrow$}} & {\textbf{T-mIoU$\uparrow$}} & {\textbf{T-SR$\uparrow$}}
& {\textbf{PSNR$\uparrow$}} & {\textbf{SSIM$\uparrow$}} & {\textbf{mIoU$\uparrow$}} & {\textbf{T-mIoU$\uparrow$}} & {\textbf{T-SR$\uparrow$}} \\
\midrule
$\dagger$ Mask2Former~\cite{mask2former}
& {-} & {-} & \underline{43.32} & 25.43 & \underline{55.96}
& {-} & {-} & \underline{41.59} & 25.15 & 40.91 \\
$\dagger$ LSeg~\cite{lseg}
& {-} & {-} & 29.17 & {-} & {-}
& {-} & {-} & 30.46 & {-} & {-} \\
$\star$ Uni3R~\cite{uni3r}
& 16.36 & 0.583 & 31.31 & {-} & {-}
& 16.74 & 0.593 & 32.17 & {-} & {-} \\
$\star$ SIU3R~\cite{siu3r}
& \underline{17.29} & \underline{0.591} & 37.38 & \underline{27.37} & 50.18
& \underline{17.82} & \underline{0.629} & 39.98 & \underline{29.39} & \underline{41.24} \\
\rowcolor{rowblue}
$\star$ S2GS (Ours)
&  \textbf{19.68} &  \textbf{0.645} &  \textbf{46.64} &  \textbf{30.19} &  \textbf{76.50}
&  \textbf{19.92} &  \textbf{0.665} &  \textbf{48.95} &  \textbf{30.01} &  \textbf{62.39} \\
\bottomrule
\end{tabular}
\end{table*}

\section{Experiments}

\subsection{Experimental Setup}

\textbf{Implementation Details.} We train and validate on ScanNet~\cite{scannet}, using the SIU3R~\cite{siu3r} preprocessed frames. The main difference lies in our long-sequence sampling strategy. 
Instead of collecting more context within a fixed viewpoint range, we construct streaming sequences by \emph{progressively extrapolating the viewpoint}: each context frame is sampled such that its viewpoint extends beyond that of the previous frame, thereby continuously expanding the viewing range over time. Detailed sequence construction, the IoU definition, and training settings are provided in the appendix.

\textbf{Baselines and Metrics.} Since there are currently no publicly available online feed-forward 3DGS methods, we compare against representative offline approaches, including SIU3R~\cite{siu3r}, Uni3R~\cite{uni3r}, and LSM~\cite{lsm}. We also include widely used 2D semantic segmentation baselines, LSeg~\cite{lseg} and Mask2Former~\cite{mask2former}. In addition, we report results for reconstruction-only Gaussian splatting baselines, including pixelSplat~\cite{pixelsplat}, MVSplat~\cite{mvsplat}, and NoPoSplat~\cite{noposplat}, which only support two-view inputs and thus are evaluated under the 2-view setting. For 3D reconstruction, we assess novel-view synthesis quality using PSNR and SSIM. For 3D scene understanding, we report per-frame semantic segmentation accuracy with mIoU, and cross-frame instance consistency using T-mIoU and T-SR. Detailed definitions of all metrics are provided in the appendix.

\begin{figure*}[t]
  \centering
  \includegraphics[width=0.8\textwidth]{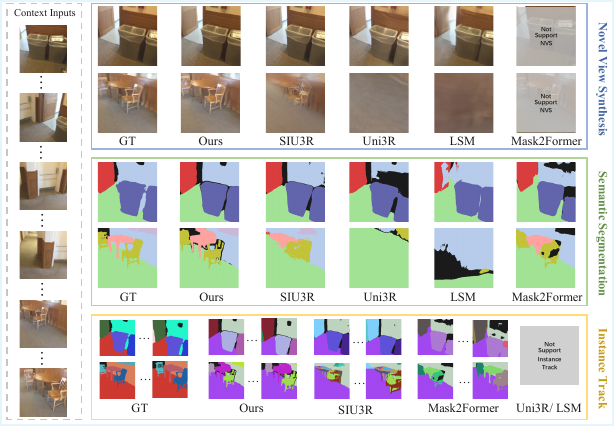}
  \caption{\textbf{Qualitative results on ScanNet dataset.}}
  \label{fig:result}
\end{figure*}

\subsection{Results.}

\textbf{Quantitative Results.} We evaluate S2GS on ScanNet and compare it with offline feed-forward baselines. As shown in Table~\ref{tab:scannet_feedforward}, under the extremely sparse 2-view setting, S2GS does not achieve the best PSNR/SSIM. This is expected, since offline baselines can exploit non-causal cross-view aggregation over the full input set to better resolve view ambiguity and occlusions when observations are highly limited. In contrast, S2GS is designed for streaming inputs and incrementally aggregates multi-view evidence as views arrive, without relying on global alignment. Consequently, it may be constrained by insufficient geometric cues at only two views. Nevertheless, as the number of input views increases (8/14/32), S2GS consistently improves and achieves strong performance in both reconstruction quality and temporal semantic/instance consistency, highlighting its effectiveness in practical streaming multi-view regimes.

\textbf{Visual comparison.} Figure~\ref{fig:result} shows qualitative ScanNet results for novel view synthesis, semantic segmentation, and instance tracking. We include two synthesis examples: one with a small (early) viewpoint/temporal gap and one with a larger (late) gap. Overall, S2GS yields sharper, more geometrically consistent renderings and more stable, temporally consistent semantic and instance predictions, while baselines degrade with larger gaps.

\begin{figure}[ht]
  \centering 
  \includegraphics[width=1.0\columnwidth]{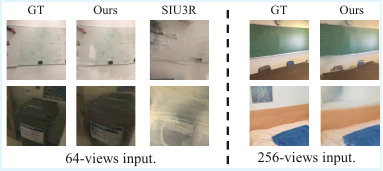}
  \caption{
    \textbf{Late-stage novel view synthesis results under longer input streams.}
  }
  \label{fig:long_nvs}
\end{figure}

\textbf{Longer-sequence performance.} Table~\ref{tab:long} and Figure~\ref{fig:long_nvs} show that for late-stage novel view synthesis, SIU3R degrades at 64 views (blur and geometric instability) and fails at 256 views, while our method preserves sharper, more consistent renderings and still produces usable results. Overall, S2GS is more efficient and scalable for long-horizon streaming.

\begin{figure}[ht]
  \centering
  \includegraphics[width=0.7\columnwidth,height=0.15\textheight,keepaspectratio=false]{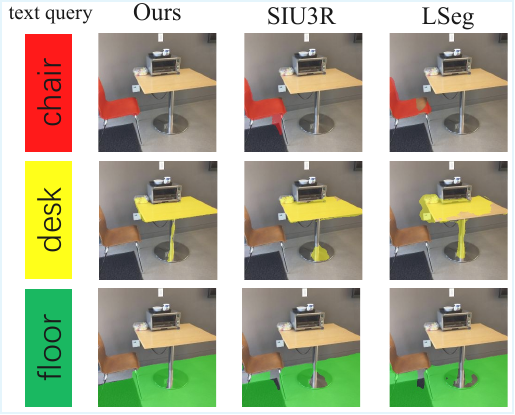}
  \caption{\textbf{Open-vocabulary language grounding with text queries.}}
  \label{fig:open_vocab}
\end{figure}

\begin{table}[t]
\centering
\caption{Comparison under Long-sequence input views.}
\label{tab:long}
\renewcommand{\arraystretch}{1.05}
\setlength{\tabcolsep}{4pt}

\definecolor{rowblue}{RGB}{230,243,255}

\begin{tabular}{c l c c c c}
\toprule
\textbf{Views} & \textbf{Method} & \textbf{PSNR$\uparrow$} & \textbf{SSIM$\uparrow$}
& \textbf{T-mIoU$\uparrow$} & \textbf{T-SR$\uparrow$} \\
\midrule

\multirow{2}{*}{64}
& SIU3R & 15.37 & 0.579 & 24.41 & 26.10 \\
& \cellcolor{rowblue}Ours
  & \cellcolor{rowblue}\textbf{18.71}
  & \cellcolor{rowblue}\textbf{0.638}
  & \cellcolor{rowblue}\textbf{26.71}
  & \cellcolor{rowblue}\textbf{45.91}\\

\midrule

\multirow{2}{*}{256}
& SIU3R & OOM & OOM & OOM & OOM \\
& \cellcolor{rowblue}Ours
  & \cellcolor{rowblue}\textbf{16.37}
  & \cellcolor{rowblue}\textbf{0.581}
  & \cellcolor{rowblue}\textbf{20.46}
  & \cellcolor{rowblue}\textbf{21.37} \\

\bottomrule
\end{tabular}
\end{table}

\begin{table}[t]
\centering
\caption{Current-frame inference time and PGM under online streaming input.}
\label{tab:system_perf}
\renewcommand{\arraystretch}{1.05}
\setlength{\tabcolsep}{2pt}
\begin{tabular}{l
S[table-format=1.3,detect-weight=true,detect-inline-weight=math]
S[table-format=1.2,detect-weight=true,detect-inline-weight=math]
S[table-format=1.3,detect-weight=true,detect-inline-weight=math]
S[table-format=2.2,detect-weight=true,detect-inline-weight=math]
S[table-format=1.5,detect-weight=true,detect-inline-weight=math]}
\toprule
\multirow{2}{*}{\textbf{Views}}
& \multicolumn{2}{c}{\textbf{SIU3R}}
& \multicolumn{2}{c}{\textbf{Ours}} \\
\cmidrule(lr){2-3}\cmidrule(lr){4-5}
& {\textbf{Time(s) $\downarrow$}} & {\textbf{PGM(GB) $\downarrow$}}
& {\textbf{Time(s) $\downarrow$}} & {\textbf{PGM(GB) $\downarrow$}} \\
\midrule
16   & 0.26  & 6.14  & 0.10  & 6.68  \\
64   & 1.52 & 42.89 & 0.12  & 9.66  \\
128  & {OOM}   & {OOM} & 0.15 & 14.48 \\
512  & {OOM}   & {OOM} & 0.35  & 41.45 \\
1024 & {OOM}   & {OOM} & 0.62  & 76.88 \\
\bottomrule
\end{tabular}
\end{table}

\begin{table}[t]
\centering
\caption{Zero-shot cross-dataset comparison under 32-view input.}
\label{tab:zero_shot_32v}
\renewcommand{\arraystretch}{1.05}
\setlength{\tabcolsep}{8pt}
\begin{tabular}{l
S[table-format=2.2,detect-weight=true,detect-inline-weight=math]
S[table-format=2.2,detect-weight=true,detect-inline-weight=math]
S[table-format=2.2,detect-weight=true,detect-inline-weight=math]
S[table-format=2.2,detect-weight=true,detect-inline-weight=math]}
\toprule
\multirow{2}{*}{\textbf{Method}}
& \multicolumn{2}{c}{\textbf{ScanNet++}}
& \multicolumn{2}{c}{\textbf{Replica}} \\
\cmidrule(lr){2-3}\cmidrule(lr){4-5}
& {\textbf{PSNR$\uparrow$}} & {\textbf{mIoU$\uparrow$}}
& {\textbf{PSNR$\uparrow$}} & {\textbf{mIoU$\uparrow$}} \\
\midrule
SIU3R & 12.85 & 33.51 & 13.14 & 21.42 \\
\rowcolor{rowblue}
S2GS & \textbf{15.33} &\textbf{41.67} &\textbf{15.66} &\textbf{37.47} \\
\bottomrule
\end{tabular}
\end{table}

\textbf{Efficiency comparison under streaming input.}
Table~\ref{tab:system_perf} reports the per-frame latency (processing the current view and updating the persistent state) and peak GPU memory (PGM) under online streaming input with different numbers of views. Our method maintains low per-frame runtime with only mild growth as the stream length increases, while SIU3R~\cite{siu3r} exhibits rapidly growing computation and memory consumption.

\begin{table}[t]
\centering
\caption{mIoU of open-vocabulary language grounding from text queries.}
\label{tab:open_vocab_miou}
\setlength{\tabcolsep}{16pt}
\renewcommand{\arraystretch}{1.1}
\begin{tabular}{lccc}
\toprule
\textbf{Method} & \textbf{Ours} & \textbf{SIU3R} & \textbf{LSeg} \\
\midrule
\textbf{mIoU} & \textbf{49.37} & 45.76 & 34.28 \\
\bottomrule
\end{tabular}
\end{table}

\textbf{Open-vocabulary language grounding results.} We further evaluate the model’s open-vocabulary language grounding capability by using free-form text queries as semantic prompts, and visualize the corresponding predicted masks in Figure~\ref{fig:open_vocab}. For text queries such as chair, desk, and floor, our method produces spatially more coherent predictions with sharper boundaries: it not only covers the main extent of the target object, but also better aligns with the geometric structure while reducing ``bleeding'' into the background or adjacent objects. Table~\ref{tab:open_vocab_miou} reports the mIoU over these text queries, where our method achieves the best performance, outperforming SIU3R and LSeg.

\textbf{Generalization evaluation.} To comprehensively assess cross-dataset generalization, we conduct zero-shot experiments on ScanNet++ \cite{scannet++} and Replica \cite{replica} : models are trained only on ScanNet and directly transferred to the target datasets under the 32-view setting. The results are reported in Table \ref{tab:zero_shot_32v}. We observe that both S2GS and the baseline SIU3R \cite{siu3r} exhibit non-trivial zero-shot generalization capability. Nevertheless, under the same training configuration, S2GS achieves better reconstruction and semantic performance on both datasets, demonstrating stronger cross-dataset generalization and robustness.

\begin{table}[t]
\centering
\caption{Ablation study on using a shared backbone vs.\ geometry--semantic decoupling.}
\label{tab:ab1}
\renewcommand{\arraystretch}{1.05}
\setlength{\tabcolsep}{12pt} 
\begin{tabular}{l
S[table-format=2.2,detect-weight=true,detect-inline-weight=math]
S[table-format=2.2,detect-weight=true,detect-inline-weight=math]
S[table-format=2.2,detect-weight=true,detect-inline-weight=math]}
\toprule
\textbf{Method} & {\textbf{mIoU$\uparrow$}} & {\textbf{T-mIoU$\uparrow$}} & {\textbf{T-SR$\uparrow$}} \\
\midrule
Shared    & 42.17 & 21.32 & 38.75 \\
\rowcolor{rowblue}
Decoupled &  \textbf{48.95} &  \textbf{30.01} &  \textbf{62.39} \\
\bottomrule
\end{tabular}
\end{table}

\begin{figure}[ht]
  \centering 
  \includegraphics[width=0.85\columnwidth]{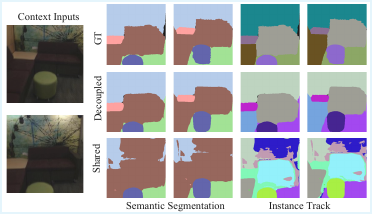}
  \caption{
    \textbf{Ablation on geometry--semantic decoupling.}
  }
  \label{figab1}
\end{figure}

\begin{table}[!t]
\centering
\caption{Ablation study on the effectiveness of query-level semantic-embedding contrastive learning.}
\label{tab:ab2}
\renewcommand{\arraystretch}{1.05}
\setlength{\tabcolsep}{13pt} 
\begin{tabular}{l
S[table-format=2.2,detect-weight=true,detect-inline-weight=math]
S[table-format=2.2,detect-weight=true,detect-inline-weight=math]
S[table-format=2.2,detect-weight=true,detect-inline-weight=math]}
\toprule
\textbf{Method} & {\textbf{mIoU$\uparrow$}} & {\textbf{T-mIoU$\uparrow$}} & {\textbf{T-SR$\uparrow$}} \\
\midrule
w/o CL & 47.13 & 28.64 & 50.13 \\
\rowcolor{rowblue}
w CL   &  \textbf{48.95} &  \textbf{30.01} &  \textbf{62.39} \\
\bottomrule
\end{tabular}
\end{table}

\begin{figure}[!tbp]
  \centering
  \includegraphics[width=0.85\columnwidth]{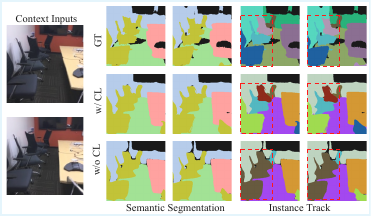}
  \caption{\textbf{Ablation on query-level contrastive learning.}}
  \label{figab2}
\end{figure}
    
\begin{table}[!t]
\centering
\caption{Ablation study on the effectiveness of distilling the base model for the geometric backbone.}
\label{tab:ab3}
\renewcommand{\arraystretch}{1.05}
\setlength{\tabcolsep}{22pt}
\begin{tabular}{l
S[table-format=2.2,detect-weight=true,detect-inline-weight=math]
S[table-format=1.3,detect-weight=true,detect-inline-weight=math]}
\toprule
\textbf{Method} & {\textbf{PSNR$\uparrow$}} & {\textbf{SSIM$\uparrow$}} \\
\midrule
w/o distill & 10.21 & 0.237 \\
\rowcolor{rowblue}
w distill   & \textbf{19.92} & \textbf{0.665} \\
\bottomrule
\end{tabular}
\end{table}

\subsection{Ablation Studies.}

\textbf{Ablation on shared backbone vs.\ geometry--semantic decoupling.}
The shared-backbone model predicts semantics using features from the geometry stream rather than extracts from 2D foundation model. As shown in Table~\ref{tab:ab1}, geometry--semantic decoupling leads to a clear improvement in per-frame semantic accuracy and yields even larger gains in temporal instance consistency metrics. Figure~\ref{figab1} further shows that the shared-backbone model tends to produce blurrier semantic segmentation, which in turn degrades instance tracking accuracy, whereas the decoupled design maintains more stable instance identities.

\textbf{Ablation on query-level semantic embedding contrastive learning.}
Table~\ref{tab:ab2} evaluates the effect of query-level contrastive learning (CL) on instance consistency under online inference. With CL, embeddings of the same instance are pulled closer while those of different instances are pushed apart, leading to a clear improvement in cross-frame instance consistency and a noticeable gain in semantic segmentation mIoU. Figure~\ref{figab2} further shows that CL helps the model better distinguish semantically similar instances within the same category in cluttered regions (red boxes).

\textbf{Ablation study on the effectiveness of distilling the base model for the geometric backbone.} In Table \ref{tab:ab3}, we evaluate the effect of the distillation loss on reconstruction quality. The results demonstrate that the distillation loss significantly improves reconstruction quality.

\section{Conclusion}
We propose S2GS, a reprocessing-free framework that incrementally maintains a persistent 3D Gaussian semantic field for online scene understanding
and reconstruction. S2GS decouples geometry and semantics: a causal geometry stream guided by geometric priors drives stable incremental reconstruction, while an independent semantic stream leverages 2D foundation features and a query-driven decoder to predict per-frame masks and identity embeddings. To reduce identity drift, we apply cross-frame contrastive alignment on query embeddings and perform lightweight online association with an instance memory. Experiments show that S2GS achieves performance on par with or better than strong offline baselines, while significantly outperforming offline global methods in scalability with sequence length.

\section*{Impact Statement}

This paper presents work whose goal is to advance the field of Machine
Learning. There are many potential societal consequences of our work, none of
which we feel must be specifically highlighted here.
\nocite{langley00}
\bibliography{example_paper}

@inproceedings{
siu3r,
title={{SIU}3R: Simultaneous Scene Understanding and 3D Reconstruction Beyond Feature Alignment},
author={Qi Xu and Dongxu Wei and Lingzhe Zhao and Wenpu Li and Zhangchi Huang and Shunping Ji and Peidong Liu},
booktitle={The Thirty-ninth Annual Conference on Neural Information Processing Systems},
year={2025}
}

@article{uni3r,
  title={Uni3r: Unified 3d reconstruction and semantic understanding via generalizable gaussian splatting from unposed multi-view images},
  author={Sun, Xiangyu and Jiang, Haoyi and Liu, Liu and Nam, Seungtae and Kang, Gyeongjin and Wang, Xinjie and Sui, Wei and Su, Zhizhong and Liu, Wenyu and Wang, Xinggang and others},
  journal={arXiv preprint arXiv:2508.03643},
  year={2025}
}

@article{uniforward,
  title={UniForward: Unified 3D Scene and Semantic Field Reconstruction via Feed-Forward Gaussian Splatting from Only Sparse-View Images},
  author={Tian, Qijian and Tan, Xin and Gong, Jingyu and Xie, Yuan and Ma, Lizhuang},
  journal={arXiv preprint arXiv:2506.09378},
  year={2025}
}

@article{iggt,
  title={IGGT: Instance-Grounded Geometry Transformer for Semantic 3D Reconstruction},
  author={Li, Hao and Zou, Zhengyu and Liu, Fangfu and Zhang, Xuanyang and Hong, Fangzhou and Cao, Yukang and Lan, Yushi and Zhang, Manyuan and Yu, Gang and Zhang, Dingwen and others},
  journal={arXiv preprint arXiv:2510.22706},
  year={2025}
}

@Article{3dgs,
      author       = {Kerbl, Bernhard and Kopanas, Georgios and Leimk{\"u}hler, Thomas and Drettakis, George},
      title        = {3D Gaussian Splatting for Real-Time Radiance Field Rendering},
      journal      = {ACM Transactions on Graphics},
      number       = {4},
      volume       = {42},
      month        = {July},
      year         = {2023}
}

@article{nerf,
  title={Nerf: Representing scenes as neural radiance fields for view synthesis},
  author={Mildenhall, Ben and Srinivasan, Pratul P and Tancik, Matthew and Barron, Jonathan T and Ramamoorthi, Ravi and Ng, Ren},
  journal={Communications of the ACM},
  volume={65},
  number={1},
  pages={99--106},
  year={2021},
  publisher={ACM New York, NY, USA}
}

@article{app1,
  title={Splat-mover: Multi-stage, open-vocabulary robotic manipulation via editable gaussian splatting},
  author={Shorinwa, Ola and Tucker, Johnathan and Smith, Aliyah and Swann, Aiden and Chen, Timothy and Firoozi, Roya and Kennedy III, Monroe and Schwager, Mac},
  journal={arXiv preprint arXiv:2405.04378},
  year={2024}
}

@inproceedings{app2,
  title={Vr-gs: A physical dynamics-aware interactive gaussian splatting system in virtual reality},
  author={Jiang, Ying and Yu, Chang and Xie, Tianyi and Li, Xuan and Feng, Yutao and Wang, Huamin and Li, Minchen and Lau, Henry and Gao, Feng and Yang, Yin and others},
  booktitle={ACM SIGGRAPH 2024 Conference Papers},
  pages={1--1},
  year={2024}
}

@article{app3,
  title={Understanding augmented reality: Concepts and applications},
  author={Craig, Alan B},
  year={2013},
  publisher={Newnes}
}

@inproceedings{pixelsplat,
  title={pixelsplat: 3d gaussian splats from image pairs for scalable generalizable 3d reconstruction},
  author={Charatan, David and Li, Sizhe Lester and Tagliasacchi, Andrea and Sitzmann, Vincent},
  booktitle={Proceedings of the IEEE/CVF conference on computer vision and pattern recognition},
  pages={19457--19467},
  year={2024}
}

@inproceedings{mvsplat,
  title={Mvsplat: Efficient 3d gaussian splatting from sparse multi-view images},
  author={Chen, Yuedong and Xu, Haofei and Zheng, Chuanxia and Zhuang, Bohan and Pollefeys, Marc and Geiger, Andreas and Cham, Tat-Jen and Cai, Jianfei},
  booktitle={European Conference on Computer Vision},
  pages={370--386},
  year={2024},
  organization={Springer}
}

@article{noposplat,
  title={No pose, no problem: Surprisingly simple 3d gaussian splats from sparse unposed images},
  author={Ye, Botao and Liu, Sifei and Xu, Haofei and Li, Xueting and Pollefeys, Marc and Yang, Ming-Hsuan and Peng, Songyou},
  journal={arXiv preprint arXiv:2410.24207},
  year={2024}
}

@article{anysplat,
  title={Anysplat: Feed-forward 3d gaussian splatting from unconstrained views},
  author={Jiang, Lihan and Mao, Yucheng and Xu, Linning and Lu, Tao and Ren, Kerui and Jin, Yichen and Xu, Xudong and Yu, Mulin and Pang, Jiangmiao and Zhao, Feng and others},
  journal={ACM Transactions on Graphics (TOG)},
  volume={44},
  number={6},
  pages={1--16},
  year={2025},
  publisher={ACM New York, NY, USA}
}

@inproceedings{dust3r,
  title={Dust3r: Geometric 3d vision made easy},
  author={Wang, Shuzhe and Leroy, Vincent and Cabon, Yohann and Chidlovskii, Boris and Revaud, Jerome},
  booktitle={Proceedings of the IEEE/CVF Conference on Computer Vision and Pattern Recognition},
  pages={20697--20709},
  year={2024}
}

@inproceedings{mast3r,
  title={Grounding image matching in 3d with mast3r},
  author={Leroy, Vincent and Cabon, Yohann and Revaud, J{\'e}r{\^o}me},
  booktitle={European Conference on Computer Vision},
  pages={71--91},
  year={2024},
  organization={Springer}
}

@inproceedings{vggt,
  title={Vggt: Visual geometry grounded transformer},
  author={Wang, Jianyuan and Chen, Minghao and Karaev, Nikita and Vedaldi, Andrea and Rupprecht, Christian and Novotny, David},
  booktitle={Proceedings of the Computer Vision and Pattern Recognition Conference},
  pages={5294--5306},
  year={2025}
}

@inproceedings{fast3r,
  title={Fast3r: Towards 3d reconstruction of 1000+ images in one forward pass},
  author={Yang, Jianing and Sax, Alexander and Liang, Kevin J and Henaff, Mikael and Tang, Hao and Cao, Ang and Chai, Joyce and Meier, Franziska and Feiszli, Matt},
  booktitle={Proceedings of the Computer Vision and Pattern Recognition Conference},
  pages={21924--21935},
  year={2025}
}

@article{streamgs,
  title={StreamGS: Online Generalizable Gaussian Splatting Reconstruction for Unposed Image Streams},
  author={Li, Yang and Wang, Jinglu and Chu, Lei and Li, Xiao and Kao, Shiu-hong and Chen, Ying-Cong and Lu, Yan},
  journal={arXiv preprint arXiv:2503.06235},
  year={2025}
}

@article{longsplat,
  title={Longsplat: Online generalizable 3d gaussian splatting from long sequence images},
  author={Huang, Guichen and Wang, Ruoyu and Gao, Xiangjun and Sun, Che and Wu, Yuwei and Gao, Shenghua and Jia, Yunde},
  journal={arXiv preprint arXiv:2507.16144},
  year={2025}
}

@article{infinitevggt,
  title={InfiniteVGGT: Visual Geometry Grounded Transformer for Endless Streams},
  author={Yuan, Shuai and Yang, Yantai and Yang, Xiaotian and Zhang, Xupeng and Zhao, Zhonghao and Zhang, Lingming and Zhang, Zhipeng},
  journal={arXiv preprint arXiv:2601.02281},
  year={2026}
}

@article{spann3r,
  title={3d reconstruction with spatial memory},
  author={Wang, Hengyi and Agapito, Lourdes},
  journal={arXiv preprint arXiv:2408.16061},
  year={2024}
}

@article{stream3r,
  title={Stream3r: Scalable sequential 3d reconstruction with causal transformer},
  author={Lan, Yushi and Luo, Yihang and Hong, Fangzhou and Zhou, Shangchen and Chen, Honghua and Lyu, Zhaoyang and Yang, Shuai and Dai, Bo and Loy, Chen Change and Pan, Xingang},
  journal={arXiv preprint arXiv:2508.10893},
  year={2025}
}

@article{streamvggt,
  title={Streaming 4d visual geometry transformer},
  author={Zhuo, Dong and Zheng, Wenzhao and Guo, Jiahe and Wu, Yuqi and Zhou, Jie and Lu, Jiwen},
  journal={arXiv preprint arXiv:2507.11539},
  year={2025}
}

@inproceedings{cut3r,
  title={Continuous 3d perception model with persistent state},
  author={Wang, Qianqian and Zhang, Yifei and Holynski, Aleksander and Efros, Alexei A and Kanazawa, Angjoo},
  booktitle={Proceedings of the Computer Vision and Pattern Recognition Conference},
  pages={10510--10522},
  year={2025}
}

@article{rnn,
  title={Recurrent neural network regularization},
  author={Zaremba, Wojciech and Sutskever, Ilya and Vinyals, Oriol},
  journal={arXiv preprint arXiv:1409.2329},
  year={2014}
}

@inproceedings{
lseg,
title={Language-driven Semantic Segmentation},
author={Boyi Li and Kilian Q Weinberger and Serge Belongie and Vladlen Koltun and Rene Ranftl},
booktitle={International Conference on Learning Representations},
year={2022}
}

@article{lsm,
  title={Large spatial model: End-to-end unposed images to semantic 3d},
  author={Fan, Zhiwen and Zhang, Jian and Cong, Wenyan and Wang, Peihao and Li, Renjie and Wen, Kairun and Zhou, Shijie and Kadambi, Achuta and Wang, Zhangyang and Xu, Danfei and others},
  journal={Advances in neural information processing systems},
  volume={37},
  pages={40212--40229},
  year={2024}
}

@inproceedings{mask2former,
  title={Masked-attention mask transformer for universal image segmentation},
  author={Cheng, Bowen and Misra, Ishan and Schwing, Alexander G and Kirillov, Alexander and Girdhar, Rohit},
  booktitle={Proceedings of the IEEE/CVF conference on computer vision and pattern recognition},
  pages={1290--1299},
  year={2022}
}

@inproceedings{scannet,
  title={Scannet: Richly-annotated 3d reconstructions of indoor scenes},
  author={Dai, Angela and Chang, Angel X and Savva, Manolis and Halber, Maciej and Funkhouser, Thomas and Nie{\ss}ner, Matthias},
  booktitle={Proceedings of the IEEE conference on computer vision and pattern recognition},
  pages={5828--5839},
  year={2017}
}

@inproceedings{scannet++,
  title={Scannet++: A high-fidelity dataset of 3d indoor scenes},
  author={Yeshwanth, Chandan and Liu, Yueh-Cheng and Nie{\ss}ner, Matthias and Dai, Angela},
  booktitle={Proceedings of the IEEE/CVF International Conference on Computer Vision},
  pages={12--22},
  year={2023}
}

@article{replica,
  title={The replica dataset: A digital replica of indoor spaces},
  author={Straub, Julian and Whelan, Thomas and Ma, Lingni and Chen, Yufan and Wijmans, Erik and Green, Simon and Engel, Jakob J and Mur-Artal, Raul and Ren, Carl and Verma, Shobhit and others},
  journal={arXiv preprint arXiv:1906.05797},
  year={2019}
}

@article{adapter,
  title={Vision transformer adapter for dense predictions},
  author={Chen, Zhe and Duan, Yuchen and Wang, Wenhai and He, Junjun and Lu, Tong and Dai, Jifeng and Qiao, Yu},
  journal={arXiv preprint arXiv:2205.08534},
  year={2022}
}

@article{siglip2,
  title={Siglip 2: Multilingual vision-language encoders with improved semantic understanding, localization, and dense features},
  author={Tschannen, Michael and Gritsenko, Alexey and Wang, Xiao and Naeem, Muhammad Ferjad and Alabdulmohsin, Ibrahim and Parthasarathy, Nikhil and Evans, Talfan and Beyer, Lucas and Xia, Ye and Mustafa, Basil and others},
  journal={arXiv preprint arXiv:2502.14786},
  year={2025}
}

@inproceedings{niceslam,
  title={Nice-slam: Neural implicit scalable encoding for slam},
  author={Zhu, Zihan and Peng, Songyou and Larsson, Viktor and Xu, Weiwei and Bao, Hujun and Cui, Zhaopeng and Oswald, Martin R and Pollefeys, Marc},
  booktitle={Proceedings of the IEEE/CVF conference on computer vision and pattern recognition},
  pages={12786--12796},
  year={2022}
}

@inproceedings{gaussianslam,
  title={Gaussian splatting slam},
  author={Matsuki, Hidenobu and Murai, Riku and Kelly, Paul HJ and Davison, Andrew J},
  booktitle={Proceedings of the IEEE/CVF Conference on Computer Vision and Pattern Recognition},
  pages={18039--18048},
  year={2024}
}

@article{omnimap,
  title={OmniMap: A General Mapping Framework Integrating Optics, Geometry, and Semantics},
  author={Deng, Yinan and Yue, Yufeng and Dou, Jianyu and Zhao, Jingyu and Wang, Jiahui and Tang, Yujie and Yang, Yi and Fu, Mengyin},
  journal={IEEE Transactions on Robotics},
  year={2025},
  publisher={IEEE}
}

@inproceedings{gs3lam,
  title={Gs3lam: Gaussian semantic splatting slam},
  author={Li, Linfei and Zhang, Lin and Wang, Zhong and Shen, Ying},
  booktitle={Proceedings of the 32nd ACM International Conference on Multimedia},
  pages={3019--3027},
  year={2024}
}

@article{vtg,
  title={VTGaussian-SLAM: RGBD SLAM for Large Scale Scenes with Splatting View-Tied 3D Gaussians},
  author={Hu, Pengchong and Han, Zhizhong},
  journal={arXiv preprint arXiv:2506.02741},
  year={2025}
}

@article{dinov2,
  title={Dinov2: Learning robust visual features without supervision},
  author={Oquab, Maxime and Darcet, Timoth{\'e}e and Moutakanni, Th{\'e}o and Vo, Huy and Szafraniec, Marc and Khalidov, Vasil and Fernandez, Pierre and Haziza, Daniel and Massa, Francisco and El-Nouby, Alaaeldin and others},
  journal={arXiv preprint arXiv:2304.07193},
  year={2023}
}

@article{hm,
  title={The Hungarian method for the assignment problem},
  author={Kuhn, Harold W},
  journal={Naval research logistics quarterly},
  volume={2},
  number={1-2},
  pages={83--97},
  year={1955},
  publisher={Wiley Online Library}
}
\bibliographystyle{icml2026}

\newpage
\appendix
\onecolumn


\end{document}